%% file: main.tex
\documentclass[10pt,twocolumn,letterpaper]{article}

\usepackage[pagenumbers]{cvpr} % To force page numbers, e.g. for an arXiv version
\usepackage{url}
\usepackage{graphicx}
\usepackage{todonotes}
\usepackage{booktabs}          % \toprule, \midrule, \bottomrule
\usepackage{graphicx}          % \resizebox
\usepackage{array}             % (optional) nicer column handling
\usepackage[table]{xcolor}     % \rowcolor, gray!25 and \arrayrulecolor
\usepackage{subcaption}    % for subtable environment
\usepackage{caption}
\usepackage{amsmath}

\usepackage{multirow}
\usepackage{algorithm}
\usepackage{wrapfig}

\usepackage[table]{xcolor}
\input{preamble}
\definecolor{cvprblue}{rgb}{0.21,0.49,0.74}
\definecolor{myred}{rgb}{0.6,0.1, 0.1}
\usepackage[pagebackref,breaklinks,colorlinks,allcolors=cvprblue]{hyperref}

\def\paperID{****} % *** Enter the Paper ID here
\def\confName{CVPR}
\def\confYear{2026}

\title{ Understanding, Accelerating, and Improving MeanFlow Training
}

\author{
{Jin-Young Kim$^1\thanks{Equal contribution}$ \quad Hyojun Go$^{2*}$ \quad Lea Bogensperger$^3$ \quad Julius Erbach$^{2,4}$ } \\
{Nikolai Kalischek$^5$ \quad Federico Tombari$^5$ \quad Konrad Schindler$^2$ \quad Dominik Narnhofer$^2$ \vspace{0.3cm}} \\
{$^1$Yonsei University \quad $^2$ETH Zurich \quad $^3$University of Zurich \quad $^4$Max Planck ETH CLS \quad $^5$Google }
}
\begin{document}
\maketitle
\input{sec/0_abstract}

\input{sec/1_intro}

\input{sec/2_related_work}

\input{sec/3_prelim}

\input{sec/4_observation}

\input{sec/5_method}

\input{sec/6_experiments}

\input{sec/7_conclusion}
{
    \small
    \bibliographystyle{ieeenat_fullname}
    \bibliography{main}
}

\input{sec/X_suppl}


\end{document}

%% file: sec/0_abstract.tex
\begin{abstract}
MeanFlow promises high-quality generative modeling in few steps, by jointly learning instantaneous and average velocity fields.
Yet, the underlying training dynamics remain unclear.
We analyze the interaction between the two velocities and find:
(i) well-established instantaneous velocity is a prerequisite for learning average velocity;
(ii) learning of instantaneous velocity benefits from average velocity when the temporal gap is small, but degrades as the gap increases; and
(iii) task-affinity analysis indicates that smooth learning of large-gap average velocities, essential for one-step generation, depends on the prior formation of accurate instantaneous and small-gap average velocities.
Guided by these observations, we design an effective training scheme that accelerates the formation of instantaneous velocity, then shifts emphasis from short- to long-interval average velocity. 
Our enhanced MeanFlow training yields faster convergence and significantly better few-step generation:
With the same DiT-XL backbone, our method reaches an impressive FID of 2.87 on 1-NFE ImageNet 256$\times$256, compared to 3.43 for the conventional MeanFlow baseline.
Alternatively, our method matches the performance of the MeanFlow baseline with 2.5$\times$ shorter training time, or with a smaller DiT-L backbone.
Our code is available at \href{https://github.com/seahl0119/ImprovedMeanFlow}{https://github.com/seahl0119/ImprovedMeanFlow}.
\end{abstract}

%% file: sec/1_intro.tex
\section{Introduction}
\label{sec:introduction}
Diffusion models~\cite{sohl2015deep, song2019generative, ho2020denoising} and Flow Matching~\cite{lipman2022flow, albergo2022building, liu2022flow} have achieved state-of-the-art results across image~\cite{wu2025qwen, batifol2025flux, ma2024sit}, video~\cite{wan2025wan, gao2025wan, krea_realtime_14b}, and 3D generation~\cite{go2025vist3a, go2025splatflow}.
However, it remains a persistent weakness that the denoising relies on many, small iteration steps, making sampling computationally expensive~\cite{hang2023efficient}. 
Higher-order samplers~\cite{song2020denoising, lu2022dpm, lu2025dpm, dockhorn2022genie, zhang2022fast, karras2022elucidating, sabour2024align} partially alleviate this, though achieving high fidelity with fewer than 10 steps remains a challenge.
Consequently, recent work has focused on models that enable inference in a few steps, or even a single step.

\input{figures/teaser}

Early approaches distill few-step generative models from pretrained multi-step diffusion models, using direct~\cite{luhman2021knowledge, zheng2023fast, salimans2022progressive}, adversarial~\cite{sauer2024adversarial, yin2024improved, sauer2024fast}, or score-based supervision~\cite{zhou2024score, luo2023diff, yin2024one}.
%
% Unlike an \emph{end-to-end} paradigm, where one- or few-step denoising is learned in a single training run, these methods follow a two-stage design: first training a conventional diffusion model, then distilling it into a one- and few-step sampler. 
%
The two-stage design increases complexity, requires two distinct training processes, and often depends on large-scale synthetic data generation~\cite{luhman2021knowledge, liu2022flow}, or on propagation through teacher–student cascades~\cite{meng2023distillation, salimans2022progressive}.

Consistency models~\cite{song2023consistency} represent a step towards one-time, end-to-end training, by enforcing consistent outputs for all samples drawn along the same denoising trajectory.
Despite various improvements~\cite{lu2024simplifying, wang2024stable, song2023improved, geng2024consistency, kim2023consistency, wang2024phased, heek2024multistep}, a substantial performance gap remains between few-step consistency models and multi-step diffusion models.
More recent research~\cite{kim2023consistency, frans2025one, boffi2024flow, zhou2025inductive, boffi2025build, sabour2025align, wang2025transition} has proposed to characterize diffusion/flow quantities along two distinct time indices. 
Among these attempts, MeanFlow~\cite{geng2025mean} stands out as a stable end-to-end training scheme that markedly narrows the gap between one-step and multi-step generation. 

The key to MeanFlow’s success is the idea of exploiting the intrinsic relationship between the \emph{instantaneous velocity} (at a single time point) and the \emph{average velocity} (integrated over a time interval), such that a single network learns both simultaneously.
However, MeanFlow training is computationally expensive, and has only been 
analyzed rather superficially. In particular, it remains poorly understood how the two coupled velocity fields interact during learning and how their interplay can be coordinated to achieve high-quality one-step generation.

Here, we investigate these learning dynamics and develop a training strategy that greatly improves both generation quality and efficiency.
Through controlled experiments, we determine that: 
(i) instantaneous velocity must be established early in the training process, because it provides the foundation for learning average velocity: if instantaneous velocities are poorly formed or corrupted, then learning average velocities fails altogether;
(ii) in the opposite direction, the time interval over which average velocities are computed (the ``temporal gap") critically determines how they impact the learning of instantaneous velocities: small gaps facilitate instantaneous velocity formation and refinement, while large gaps destabilize it;
(iii) task affinity analysis reveals that one should initially focus on small-gap average velocities, which lay the foundation for learning the large-gap average velocities that are required for one-step generation.

Standard MeanFlow training ignores these subtle, but impactful dynamics. Throughout the training process, it applies the same, fixed loss function and sampling scheme, disregarding the complex dependencies between the two velocity fields.
%simply enforces fixed losses on both instantaneous velocity and large-gap average velocity.
%
This naive training objective interferes with the early formation of reasonable instantaneous velocities, which in turn delays the learning of average velocities. Ultimately, the current training practice significantly degrades overall performance compared to what would be achievable with a given model and dataset, and also slows down the training.

To remedy these issues, we propose a simple yet effective extension of MeanFlow training. 
To quickly establish reasonable instantaneous velocities, we adopt acceleration techniques from diffusion training~\cite{choi2022perception,hang2023efficient, go2023addressing, wang2025closer, kim2024denoising}.
To support the learning of correct average velocities, we design a progressive weighting scheme.
In early training stages, the weighting prioritizes small gaps, which reinforces instantaneous velocity formation and prepares the ground for large-gap learning.
As training progresses, the weighting gradually transitions to equal weighting across all gap lengths, ensuring accurate average velocities over large gaps, which are the vital ingredient for few-step inference.

Empirically, the enhanced training protocol substantially improves the generation results and also accelerates convergence.
On the standard 1-NFE ImageNet~\cite{deng2009imagenet} $256\times256$ benchmark, we improve the FID of MeanFlow-XL from FID 3.43 to 2.87, see Fig.~\ref{fig:teaser}. To reach the performance of conventional training, our improved training scheme needs 2.5$\times$ fewer iterations. Remarkably, it is even capable of matching that same performance with a smaller DiT-L backbone.

In a broader context, our work shows that there is still a lot of untapped potential to accelerate recent few-step generative models. With a better understanding of their internal dynamics and numerical properties, high-quality real-time generation may well be achievable.

% On the standard 1-NFE ImageNet~\cite{deng2009imagenet} $256\times256$ benchmark (Fig.~\ref{fig:teaser}), DiT-L trained with our method achieves an FID of 3.47, being comparable with DiT-XL with standard MeanFlow (FID 3.84). 
% On the DiT-XL backbone, our approach reaches an FID of 2.87 versus 3.43 with standard MeanFlow, surpassing the baseline by $17.3\%$ in less than half the training epochs.

%% file: figures/teaser.tex
\begin{figure}
    \centering
    \begin{subfigure}[t]{\linewidth}
        \centering
        \includegraphics[width=\linewidth]{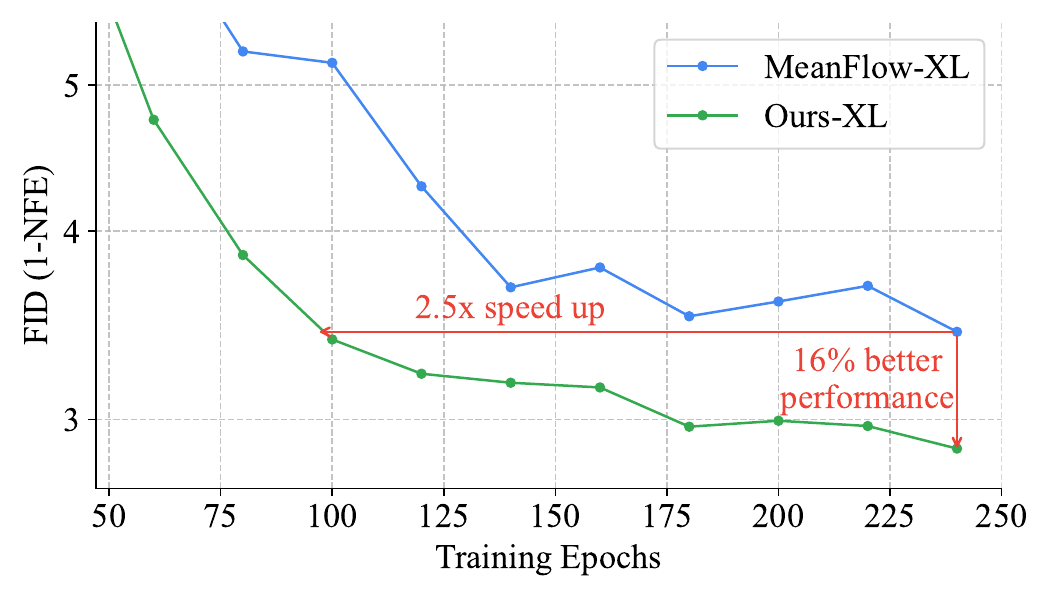}
        \vspace{-1.8cm}
    \end{subfigure}
    \begin{subfigure}[t]{\linewidth}
        \centering
        \includegraphics[width=\linewidth]{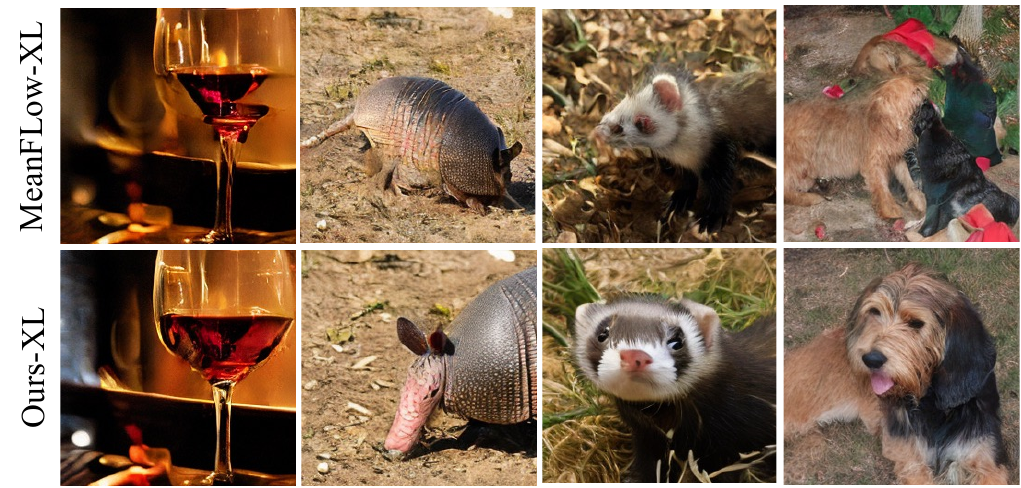}
    \end{subfigure}
    \vspace{-0.7cm}
    \caption{
    Our novel, enhanced training strategy reaches the performance of MeanFlow-XL in $\approx$2.5$\times$ fewer training epochs, and converges to a final model with superior performance ($\approx$16\% lower FID).
    }
    \label{fig:teaser}
    \vspace{-0.3cm}
\end{figure}

%% file: sec/2_related_work.tex
\section{Related Work}
\label{sec:related_work}

% \ks{in bib-file, check and fix arxiv entries that have in the meantime been accepted / published. Otherwise it looks like a careless resubmission.} -> Done!

\paragraph{Acceleration of diffusion and flow matching training.}
Diffusion models~\cite{sohl2015deep, song2019generative, ho2020denoising} gradually perturb data with noise and train a network to reconstruct the clean signal.
This noise-injection and denoising process can be described via stochastic differential equations (SDEs) or, equivalently, as a probability-flow ordinary differential equation (ODE)~\cite{karras2022elucidating, song2020score}.
Flow Matching~\cite{lipman2022flow, albergo2022building, liu2022flow} extends this with velocity fields, enabling the model to learn transport paths connecting data and reference distributions.

Training these models is computationally expensive~\cite{ramesh2022hierarchical}. 
To accelerate it, prior work mostly focuses on critical timesteps using two strategies: (1) timestep-dependent loss weighting based on SNR~\cite{hang2023efficient}, perceptual quality~\cite{choi2022perception}, or uncertainty~\cite{go2023addressing, karras2024analyzing}, and (2) modified sampling distributions~\cite{zheng2024non, zheng2024beta,zheng2025bidirectional}.
Furthermore, there are hybrid approaches that combine the two~\cite{wang2025closer}, as well as adaptive schedulers~\cite{kim2024denoising}.

We will leverage some of the existing acceleration techniques to speed up the formation of instantaneous velocity, which is important to speed up convergence, see~\cref{sec:observation}.

\vspace{-0.4cm}
\paragraph{Few-step generative models.}
% Early work distills a pretrained many-step diffusion model into a few-step student via simple distillation~\cite{luhman2021knowledge} or progressive distillation~\cite{salimans2022progressive, berthelot2023tract}. 
% While effective, several early approaches relied on generating large synthetic datasets, motivating data-free distillation methods~\cite{gu2023boot, meng2023distillation, zhou2024guided}. 
% Beyond direct output matching, alternative supervision signals have been explored, including adversarial objectives~\cite{sauer2024adversarial, yin2024improved, sauer2024fast}, score-based distillation~\cite{zhou2024score, luo2023diff, yin2024one}, moment matching~\cite{salimans2024multistep}, operator learning~\cite{zheng2023fast}, and physics-informed losses~\cite{tee2024physics}. 

Early work on few-step models revolves around distillation~\cite{luhman2021knowledge}, progressive~\cite{salimans2022progressive, berthelot2023tract}, data-free approaches~\cite{gu2023boot, meng2023distillation, zhou2024guided}, and various forms of alternative supervision,\eg adversarial~\cite{sauer2024adversarial, yin2024improved, sauer2024fast}, score-based~\cite{zhou2024score, luo2023diff, yin2024one}, moment matching~\cite{salimans2024multistep}, operator learning~\cite{zheng2023fast}, and physics-informed losses~\cite{tee2024physics}.
These are not end-to-end pipelines; they are two-stage (or tightly scheduled) procedures that require manually selected transition points—\ie, deciding when to stop teacher training and begin distillation~\cite{frans2025one}.

Consistency models~\cite{song2023consistency} pioneered a major advance, enabling \emph{end-to-end} training of few-step generators, and they have also been used in distillation setups~\cite{li2024connecting, zhang2025improving, liu2025see}.
The core principle is to enforce that model outputs from any two points along the same trajectory are identical.
This foundational concept has inspired a wide range of follow-ups, such as improving training stability and simplicity~\cite{lu2024simplifying, wang2024stable, song2023improved, geng2024consistency}, extending the framework to multi-step~\cite{kim2023consistency, wang2024phased, heek2024multistep}, adapting to latent space models~\cite{luo2023latent}, and incorporating adversarial loss~\cite{kim2023consistency, kong2024act}.
Despite these advances, a substantial gap remains between the few-step performance of end-to-end training and the performance of fully multi-step diffusion models.

More recently, several works have proposed to learn diffusion and flow quantities between two time points~\cite{kim2023consistency, frans2025one, boffi2024flow, zhou2025inductive, boffi2025build, sabour2025align, wang2025transition}.
For example, Flow Maps~\cite{boffi2024flow} define the integral of the flow between time points and learn it via matching losses. 
Shortcut Models~\cite{frans2025one} augment flow matching with a regularization loss to learn shortcut paths, and Inductive Moment Matching (IMM)~\cite{zhou2025inductive} enforces self-consistency of stochastic interpolants across time.

Among the methods developed so far, MeanFlow~\cite{geng2025mean} stands out: it substantially narrows the performance gap between few-step and full multi-step diffusion models, while being trained end-to-end.
The aim of the present paper is to understand, improve, and accelerate the training of MeanFlow.
There are a few concurrent efforts:
AlphaFlow~\cite{zhang2025alphaflow} replaces the original MeanFlow objective with a softened one. In contrast, we retain the MeanFlow formulation and improve it based on a careful empirical analysis.
CMT~\cite{hu2025cmt} splits learning into multiple stages, whereas our method preserves the simplicity of end-to-end MeanFlow training.

% Concurrent works
% - ALPHAFLOW: UNDERSTANDING AND IMPROVING MEANFLOW MODE
% - CMT: MID-TRAINING FOR EFFICIENT LEARNING OF CONSISTENCY, MEAN FLOW, AND FLOW MAP MODELS

%% file: sec/3_prelim.tex
\section{Background: MeanFlow}
\label{sec:background}

\paragraph{Flow matching.}
Flow matching~\cite{lipman2022flow,liu2022flow,albergo2022building} learns a time-dependent vector field $ v_\theta(z_t, t)$ that transports a (typically Gaussian) source distribution $\epsilon \sim p_1(\epsilon)$ to a target data distribution $x \sim p_0(x)$. This transformation is defined as the solution to the ODE that characterizes the flow $\Phi$
%\vspace{-0.3cm}
\begin{equation}
    \tfrac{d}{dt}\Phi_t(z) = v_t(\Phi_t(z)).
    \label{eq:flow_ode}
% \vspace{-0.1cm}
\end{equation}
A valid velocity field can be learned by optimizing the conditional flow matching objective, which utilizes a tractable conditional velocity $v_t(z_t|\epsilon)$ instead of the intractable marginal velocity $v_t(z_t)$~\cite{lipman2022flow}.
Different choices for the conditional flow paths are possible, arguably, the simplest and most popular ones are \textit{optimal transport conditional} velocity fields. 
Given a pair $(x,\epsilon)$ and a time $t\in[0,1]$, $z_t$ is the linear interpolant $z_t=(1-t)\,x+t\,\epsilon$, whose conditional velocity~\cite{lipman2022flow} is the time derivative $v_t(z_t|\epsilon) = \dot z_t= \epsilon - x$.
The neural network $v_\theta$ is by minimizing the conditional flow-matching loss $\mathcal{L}_{\mathrm{CFM}} = \mathbb{E}_{x, \epsilon, t} [
\left\|
v_\theta(z_t, t) - v_t(z_t\mid\epsilon)
\right\|_2^2]$.
Data samples are generated by solving the probability-flow ODE in Eq.~\ref{eq:flow_ode}.

\vspace{-0.4cm}
\paragraph{MeanFlow.}
The core idea of MeanFlow~\cite{geng2025mean} is to interpret the flow-matching velocity as the \textbf{instantaneous velocity} ($v$) at time $t$ and to learn an \textbf{average velocity} ($u$) between time points $r$ and $t$ defined by the MeanFlow identity
\vspace{-0.15cm}
\begin{equation}
    u(z_t, r, t) \triangleq \frac{1}{t-r} \int_{r}^{t} v_t\!\left(z_\tau, \tau\right)\, d\tau.
\vspace{-0.05cm}
\end{equation}
By learning $u$ with a neural network $u_\theta$, MeanFlow approximates the finite-time ODE integral in Eq.~\ref{eq:flow_ode}, enabling a single update step $ z_r\;=\;z_t-(t-r)\,u_\theta(z_t,r,t)$ that replaces multiple small solver steps.
To train $u_\theta$, MeanFlow exploits the relationship between $u$ and $v$, which is defined as:
\vspace{-0.05cm}
\begin{equation}
    u(z_t,t,r) = v_t(z_t,t) - (t-r) (v_t(z_t,t)\partial_x u_\theta + \partial_t u_\theta).
\vspace{-0.05cm}
\end{equation}
The overall objective is then given by:
\vspace{-0.05cm}
\begin{equation}
\label{eq:loss}
    \mathcal{L}_\text{MF} = \mathbb{E}_{x, \epsilon, t, r}[\| u_\theta(z_t, r, t) - \text{sg}(u_\text{tgt})\|_2^2], 
\vspace{-0.05cm}
\end{equation}
where $u_{\mathrm{tgt}} = v_t(z_t,t) -(t-r)(v_t(z_t,t)\partial_z u_\theta + \partial_t u_\theta)$ and $\mathrm{sg}(\cdot)$ denotes the stop-gradient operation.

The training strategy proposed by~\cite{geng2025mean} samples a portion of the minibatch with $t=r$, in which case 
the MeanFlow objective in Eq.~\ref{eq:loss} reduces to flow matching by learning the instantaneous velocity $v$. Hence, their loss can be interpreted as the sum of two terms for the average velocity $\mathcal{L}_u(z_t, r, t)$ and the instantaneous velocity $\mathcal{L}_v(z_t, t)$:
\begin{equation}
    \label{eq:meanflow_loss_decompose}
    % \mathcal{L}_\text{all} = \mathcal{L}_\text{MF}(z_t, r, t) + \lambda_v~\mathcal{L}_v(z_t, t).
    \mathcal{L}_\text{MF}=\mathbb{E}_{x, \epsilon, t, r}\left[\mathcal{L}_u(z_t,r,t)\!\cdot\!\mathbb{I}(t\!\neq\!r)+\mathcal{L}_v(z_t,t)\!\cdot\!\mathbb{I}(t\!=\!r)\right].
\vspace{-1mm}
\end{equation}

% $\mathcal{L}_u(\cdot, r, t)$ $\mathcal{L}_v(\cdot, t)$

% $\mathcal{L}_v(z_t, t)$, $\mathcal{L}_u(z_t, r, t)$ 

%% file: sec/4_observation.tex
\section{Observations}
\label{sec:observation}

As described in Section~\ref{sec:background}, the MeanFlow objective decomposes into learning instantaneous velocity $v$ and average velocity $u$. 
In the following, we study how these coupled quantities interact during training and how to optimize their interplay to maximize one-step generation performance.
For these experiments, we use a DiT-B/4~\cite{peebles2023scalable} architecture and the ImageNet $256\times 256$ dataset~\cite{deng2009imagenet}. 
Further details are provided in Appendix~\textcolor{cvprblue}{A}.

\subsection{Impact of \texorpdfstring{$v$}{v}-Learning on \texorpdfstring{$u$}{u}-Learning}
\label{sec:obs_v_impact_on_u}
We first investigate how learning the \emph{instantaneous velocity} $v$ affects the quality of the learned \emph{average velocity} $u$.
Through controlled experiments, we demonstrate that establishing $v$ is a prerequisite for $u$-learning. 
%Intuitively, integrating unestablished $v$ for learning $u$ leads to unstable and poor convergence, highlighting the importance of solid $v$ formation early in training.

\vspace{-0.4cm}
\paragraph{Learning $v$ facilitates learning $u$.}
\input{figures/obs1_clue1}
We examine whether and how instantaneous velocity $v$ affects the learning of the average velocity $u$.
We conduct a two-stage training: we first train the model with $v$-loss ($\mathcal{L}_v(z_t, t)$ in Eq.~\ref{eq:meanflow_loss_decompose}) and then finetune it with $u$-loss ($\mathcal{L}_u(z_t, r, t)$ in Eq.~\ref{eq:meanflow_loss_decompose}).
This experimental setup makes it possible to analyze what effect $v$-pretraining has on learning $u$. 
We use 1-NFE FID as an evaluation metric for the quality of $u$.

Figure~\ref{fig:observation_1} shows the results of two complementary settings. 
In the first setting (top), we fix the budget for finetuning $u$ at 60 epochs while varying the duration of $v$-pretraining between \{0, 5, 10, 15, 20\} epochs.
In the second setting (bottom), we fix the total training budget at 80 epochs and allocate \{0, 5, 10, 15, 20\} epochs to $v$-pretraining, with the remainder dedicated to $u$-finetuning. 
Clearly, investing more into $v$-pretraining yields more stable and accurate $u$-learning.
In the second setting, even under a fixed compute budget, prioritizing $v$ earlier is more effective and accelerates convergence.
Overall, the results suggest that a well-formed $v$ is a necessary prerequisite for subsequent $u$-learning.

\vspace{-0.4cm}

\paragraph{Corruption in $v$-learning disrupts $u$-learning.}
\input{figures/obs1_clue2}
To examine the opposite case, we investigate whether $u$ can be accurately learned when $v$-learning is deliberately corrupted.
During MeanFlow training, we intentionally inject Gaussian noise (scaled by $k\!\cdot\!\|v_t(z_t | \epsilon)\|$) into the target conditional velocity of the $v$-loss, thereby degrading $v$-learning while leaving the $u$-loss intact.
We again use 1-NFE FID as an evaluation metric for $u$-learning quality and illustrate the results across noise scales $k$ in Fig.~\ref{fig:observation_2}.
Even with small noise ($k = 0.03$), $u$-learning severely degrades.
In other words, a corrupted instantaneous velocity makes learning of the average velocity a lot harder.

\vspace{-0.4cm}
\paragraph{Implication.}
The two experiments above reveal symmetric dependencies: $u$-learning benefits from well-formed $v$, while failing when $v$ is corrupted.
This aligns with the mathematical structure of MeanFlow---$u$ is defined as the temporal integral of $v$, so the latter must be reasonably well established to learn the former.
These findings suggest that instantaneous velocity should be prioritized early in training, to lay the foundation for subsequent learning of the average velocity.

% \paragraph{Implications.}
% The two experiments jointly demonstrate that establishing an accurate instantaneous velocity field ($v$) is a necessary foundation for effective average-velocity ($u$) learning.  
% Pretraining on $\mathcal{L}_v$ provides a stable local velocity basis that allows the model to integrate reliably over time, leading to faster and more consistent convergence once $\mathcal{L}_u$ supervision is introduced.  
% Conversely, when $v$-learning is corrupted or underdeveloped, $u$-learning deteriorates sharply, confirming that the quality of $u$ is inherently constrained by the accuracy of its underlying $v$.  
% These findings highlight a hierarchical dependency between $v$ and $u$: the instantaneous velocity defines the local dynamics that the average velocity must integrate, and stable large-gap behavior cannot emerge without first forming coherent short-gap dynamics.  
% This interdependence motivates training schemes that explicitly align or warm up $v$ before emphasizing $u$, ensuring more efficient and robust mean-flow learning.

\subsection{Impact of \texorpdfstring{$u$}{u}-Learning on \texorpdfstring{$v$}{v}-Learning}
\label{sec:obs_u_impact_on_v}
\input{figures/obs2_clue1}
\input{figures/obs2_clue1_2}

Complementary to~\cref{sec:obs_v_impact_on_u}, we analyze how the temporal gap $\Delta t = t - r$ in average velocity supervision influences the instantaneous velocity.
We consider two initializations: (1) a model pretrained with $v$-loss for 40 epochs, and (2) random initialization.
Both models are then finetuned with the $u$-loss for 40 epochs while restricting $\Delta t$ to one of four ranges: $[0.1, 0.3]$, $[0.3, 0.5]$, $[0.5, 0.7]$, or $[0.7, 0.9]$.
The experimental design exposes how $u$-supervision with different $\Delta t$ modifies a pretrained instantaneous velocity, and how effectively it forms one from scratch.
To measure the quality of $v$, we set $v(z_t, t) = u_\theta(z_t, t, t)$ and evaluate 32-NFE FID, see results in~\cref{fig:observation_3}.

\vspace{-0.4cm}
\paragraph{Small $\Delta t$ supervision forms $v$.}
Restricting $u$-learning to small temporal gaps ($\Delta t \in [0.1, 0.3]$) reveals two notable effects (Fig.~\ref{fig:observation_3}).
First, a model trained from scratch with this $u$-loss achieves a 32-NFE FID, comparable to 40 epochs of $v$-pretraining (green line), demonstrating that small-$\Delta t$ supervision is a viable proxy for $v$-learning.
Second, when finetuning the $v$-pretrained model, the same $u$-loss yields additional FID gains, indicating that it also improves the pretrained instantaneous velocity.
Together, these results demonstrate that small-$\Delta t$ supervision provides an effective learning signal for the instantaneous velocity $v$.

\vspace{-0.4cm}
\paragraph{Large $\Delta t$ supervision deteriorates $v$.}
In contrast, $u$-learning with larger temporal gaps ($\Delta t \in [0.3, 0.5]$, $[0.5, 0.7]$, $[0.7, 0.9]$) yields poor 32-NFE FID, with both initializations. 
\ie, large-$\Delta t$ supervision is of limited use to construct $v$ from scratch and, even worse, seriously degrades an already pretrained instantaneous velocity.

\vspace{-0.4cm}
\paragraph{Implication.}
When training is not properly managed, there is a self-destructive dynamic: $u$-learning requires a stable $v$ foundation, yet large-$\Delta t$ supervision destabilizes it.
Together with the earlier finding that $v$ should be established early (Sec.~\ref{sec:obs_v_impact_on_u}), this translates into the following guideline: suppress large-$\Delta t$ supervision in early training stages, to avoid disrupting $v$-learning.

\subsection{Task Affinity Analysis}
From Sec.~\ref{sec:obs_v_impact_on_u}, we observe that $v$ should be established early as a foundation for $u$-learning.
However, as shown in Sec.~\ref{sec:obs_u_impact_on_v}, supervision with large temporal gaps destabilizes $v$, while supervision with small temporal gaps can actually benefit $v$ learning.
Therefore, we should exclude large-$\Delta t$ supervision from the early training phase, which should be dedicated to forming $v$.
In this regard, two viable strategies emerge: (1) pure $v$-learning via $v$-loss, or (2) combining $u$-loss with small temporal gap supervision, which serves as an effective signal for $v$-learning.

To determine whether pure $v$-loss or small-$\Delta t$ $u$-loss better prepares the ground for large-$\Delta t$ learning, we resort to the Task Affinity Score (TAS)~\cite{fifty2021efficiently, go2023addressing, standley2020tasks}.
TAS measures how smoothly two tasks can be jointly trained, quantifying their lack of conflict through training iterations.
We compute this between $v$-loss and $u$-loss (across different $\Delta t$-ranges) under three distinct initialization schemes:
\emph{baseline}: random initialization;
\emph{strategy 1}: pretrained with $v$-loss for 40 epochs;
\emph{strategy 2}: pretrained with $u$-loss ($\Delta t \in [0.1, 0.3]$) for 40 epochs.
The results, shown in~\cref{fig:observation_4}, clarify how to best initialize large-$\Delta t$ learning.

The models pretrained with strategies 1 and 2 both have higher TAS than random initialization across all temporal gap ranges, showing that both pretraining strategies create a more favorable regime for joint MeanFlow learning.
Among the two, pretraining with small-$\Delta t$ supervision exhibits a stronger affinity for large-$\Delta t$ regimes compared to pure $v$-loss pretraining, indicating that small-$\Delta t$ supervision provides a more favorable initialization for the later learning stages that extend $u$ to large temporal gaps.

\vspace{-0.4cm}
\paragraph{Implication.}
Small-$\Delta t$ supervision (Strategy 2) creates a better initialization for the challenging large-$\Delta t$ regime than pure $v$-loss (Strategy 1). Hence, small-$\Delta t$ supervision should be included early in training, preparing for the subsequent stages that learns $u$ over large temporal gaps---the mode ultimately required for one-step generation.

%% file: figures/obs1_clue1.tex
\begin{figure}[t]
    \centering
    % lenend figure
    \begin{subfigure}[t]{0.9\linewidth}
        \centering
        \includegraphics[width=\linewidth]{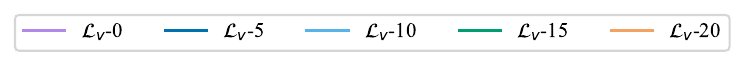}
        % \caption{\textcolor{blue}{HJ: is subfigure caption needed?}}
    \end{subfigure}
    \vspace{-5pt}
    \begin{subfigure}[t]{0.97\linewidth}
        \centering
        \includegraphics[width=\linewidth]{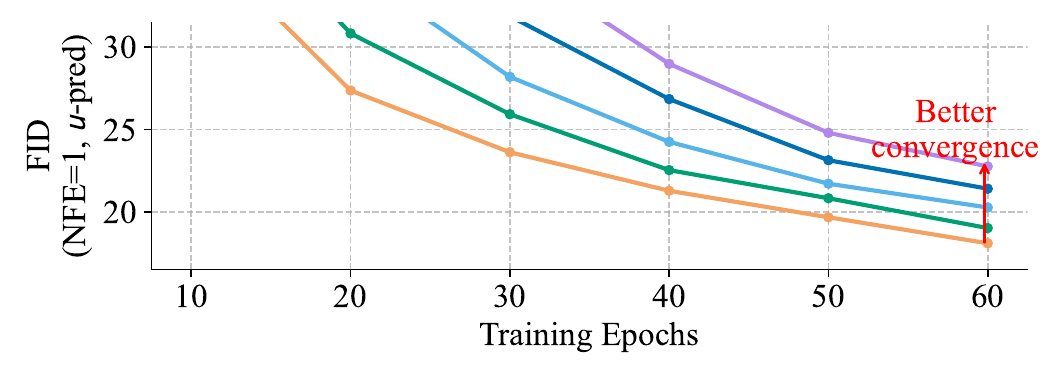}
        % \caption{\textcolor{blue}{HJ: is subfigure caption needed?}}
    \end{subfigure}
    \vspace{-5pt}
    \begin{subfigure}[t]{0.99\linewidth}
        \centering
        \includegraphics[width=\linewidth]{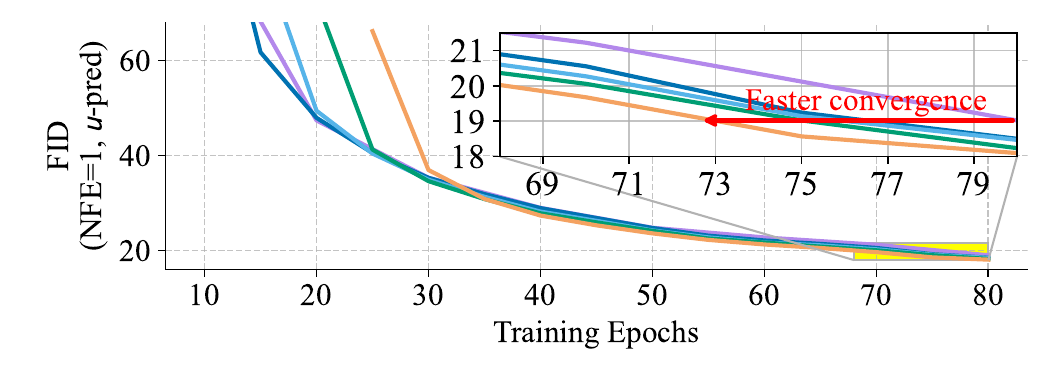}
        % \caption{\textcolor{blue}{HJ: is subfigure caption needed?}}
    \end{subfigure}
    \vspace{-0.2cm}
    \caption{\textbf{$v$-learning facilitates $u$-learning.} \emph{(Top)} 1-NFE FID during $u$-finetuning according to $v$-pretraining epochs.
    \emph{(Bottom)} 1-NFE FID under a fixed 80-epoch budget with varying allocation between $v$-pretraining and $u$-finetuning.
    Both settings show that investing in $v$-learning improves $u$-learning quality.
    }
    \vspace{-0.2cm}
    \label{fig:observation_1}
\end{figure}

%% file: figures/obs1_clue2.tex
\begin{figure}
    \centering
    \includegraphics[width=\linewidth]{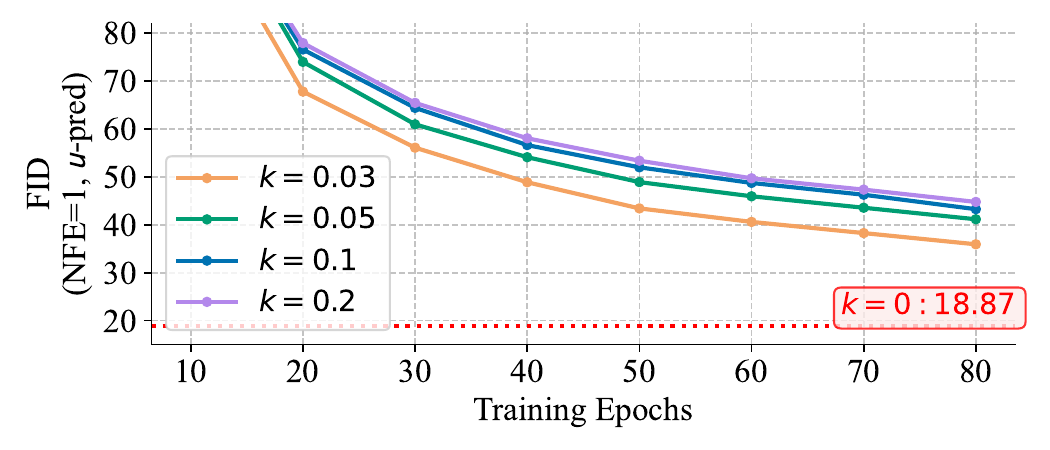}
    \vspace{-0.6cm}
    \caption{
    \textbf{Corruption in $v$-learning disrupts $u$-learning.} 
    1-NFE FID when training with $\mathcal{L}_{\mathrm{MF}}$ while injecting Gaussian noise scaled by $k\!\cdot\!\|v_t(z_t|\epsilon)\|$ into the target velocity of $\mathcal{L}_v$. 
    Even small noise ($k = 0.03$) disrupts $v$-learning and severely degrades $u$-learning performance compared to clean training ($k=0$).
    }
    \vspace{-0.2cm}
    \label{fig:observation_2}
\end{figure}

    % curvature
    % decomposing sensitive analysis by using curvature and magnitude.
    % decomposing timestep.

%% file: figures/obs2_clue1.tex
\begin{figure}
    \centering
    \includegraphics[width=1\linewidth]{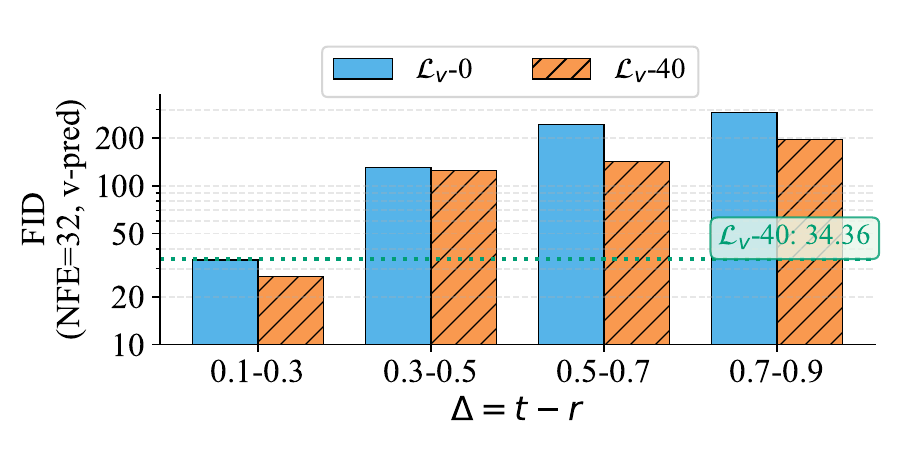}
    \vspace{-0.9cm}
    \caption{
    \textbf{Impact of $\Delta t$ of $u$-learning on $v$-learning.}
    32-NFE FID after 40 epochs of $u$ finetuning across different $\Delta t$ ranges, starting from either random initialization (blue) or $v$-pretrained model (orange, 40 epochs).
    Small $\Delta t$ enables constructing and improving $v$, while large $\Delta t$ degrades pretrained $v$. The green line denotes the performance of the $v$-pretrained model.
    }
    \vspace{-0.2cm}
    \label{fig:observation_3}
\end{figure}

%% file: figures/obs2_clue1_2.tex
\begin{figure}
    \centering
    \includegraphics[width=1\linewidth]{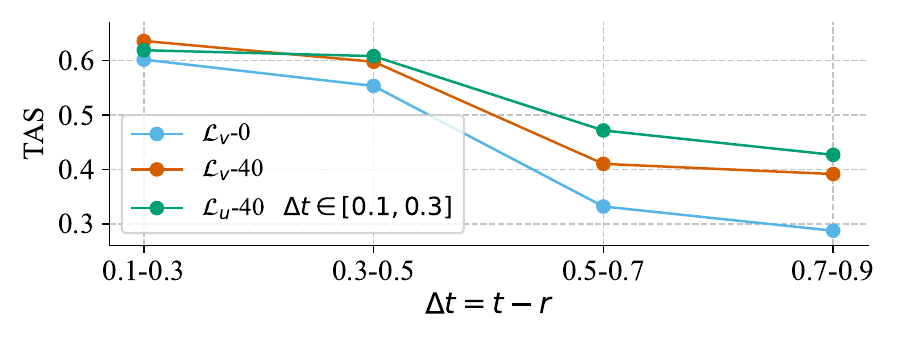}
    \vspace{-0.9cm}
    \caption{
    \textbf{Task affinity between $v$- and $u$-learning across $\Delta t$ ranges.}
    Small-$\Delta t$ $u$-pretraining achieves higher affinity for large $\Delta t$ compared to $v$-pretraining, providing a better regime for learning large-gap average velocity with instantaneous velocity.
    }
    \vspace{-0.2cm}
    \label{fig:observation_4}
\end{figure}

%% file: sec/5_method.tex
\section{Method}
\label{sec:method}

In Sec.~\ref{sec:observation}, we made three key observations: 
\textbf{(O1)} instantaneous velocity ($v$) must be established early, as it provides the foundation for learning average velocity ($u$)—when $v$ is poorly formed or corrupted, $u$-learning fails; 
\textbf{(O2)} the temporal gap determines how $u$-learning affects $v$: small temporal gaps facilitate formation and refinement of $v$, while large temporal gaps degrade it; 
and \textbf{(O3)} task affinity analysis reveals that including small-gap supervision creates a more favorable initialization for learning the large-gap average velocity required for one-step generation.

The original MeanFlow objective does not consider these properties and thus trains with standard $v$-loss and the whole range of $\Delta t$ from the start.
Thus, the training suffers from the observed inefficiencies: it fails to rapidly form $v$, which in turn delays learning of $u$, since it depends on well-formed instantaneous velocities; resulting in slow training and suboptimal model performance.

We translate our insights into a unified training strategy that directly addresses these limitations, thereby accelerating convergence and improving overall generation quality.
Our strategy has two components: (1) the use of well-established training acceleration techniques for diffusion and flow models to rapidly form $v$, and (2) progressive weighting of the $u$-loss ($\mathcal{L}_u(z_t, r, t)$), such that it prioritizes small-$\Delta t$ early in training and gradually transitions to uniform weighting across all $\Delta t$.

\vspace{-0.4cm}
\paragraph{Accelerating $v$-learning.}
To rapidly establish $v$, we adopt two standard acceleration techniques: specialized timestep sampling and time-dependent loss weighting.
For timestep sampling, we replace the base sampling in Eq.~\ref{eq:loss} with a custom distribution $p_{\text{acc}}(t)$.
For loss weighting, we apply a time-dependent weight $\alpha(t)$ to the $v$-loss, modifying the corresponding term in Eq.~\ref{eq:loss} to $\alpha(t) \cdot \mathcal{L}_v(z_t, t)$.
The specific forms of $p_{\text{acc}}(t)$ and $\alpha(t)$ are adopted from established acceleration methods~\cite{choi2022perception, go2023addressing, karras2024analyzing, hang2023efficient, zheng2024non, zheng2024beta, zheng2025bidirectional, wang2025closer} and are designed to focus the model training on more critical timesteps, thereby accelerating convergence.

\input{tables/imagenet_benchmark}

\vspace{-0.4cm}
\paragraph{Progressive $\mathcal{L}_u$ weighting.}
We weight $\mathcal{L}_u(z_t, r, t)$ to emphasize small temporal gaps early in training, then gradually transition to uniform weighting.
Specifically, we use
\[
\beta(\Delta t, s) = 1 - s + \lambda s (1 - \Delta t),
\]
%where $s \in [0,1]$ is the training progress (decreasing from 1 at start to 0 at end). 
where $s\!\in\![0,1]$ denotes the training progress.
At initialization $s=1$ and $\beta(\Delta t,1) = \lambda (1- \Delta t)$ prioritizes small $\Delta t$; at convergence $\beta(\Delta t, 0) = 1$ weights all gaps equally.
To maintain a uniform expectation at initialization, we set $\lambda = 1 / \mathbb{E}_{\Delta t}[1-\Delta t]$.
We use a simple linear schedule $s = 1 - i/T$, where $i$ and $T$ denote the current iteration and the total number of iterations, respectively. 
%We parameterize the schedule $s$ using the current iteration $i$ and the total iterations $T$. 
%A linear schedule uses $s = 1 - i/T$. 
%For slower-then-faster transitions, we can set $s = 1 - (i/T)^k$ with $k > 1$; conversely, $k < 1$ can yield faster-then-slower decay.
A slower initial transition can be achieved by setting $s = 1 - (i/T)^k$ with $k > 1$; conversely, $k < 1$ yields a faster transition.

\vspace{-0.4cm}
\paragraph{Integration with MeanFlow components.}
MeanFlow training employs a number of stabilization techniques, including specialized loss metrics and sampling strategies. In Appendix~\textcolor{cvprblue}{B} we provide detailed instructions on how to integrate our proposed adaptations with these components.

%% file: tables/imagenet_benchmark.tex
\begin{table*}[t]
\centering
\begin{minipage}[t]{0.54\textwidth}
\centering
\resizebox{0.99\linewidth}{!}{
\begin{tabular}{lcccc}
\toprule
\multirow{2}{*}{Method} & \multirow{2}{*}{\#Params} & \multirow{2}{*}{Epoch} & 1-NFE & 2-NFE\\
 &   &  & FID$\downarrow$ & FID$\downarrow$ \\
\cmidrule{1-5}
\multicolumn{5}{l}{\textit{\textbf{Comparison to MeanFlow across model sizes}}} \\ \addlinespace[2pt]
\; MeanFlow-B/4 & 131M & 240 & 11.58 & 7.85\\
\rowcolor{gray!25} \; \quad + Ours w MinSNR  & 131M & 240 & \textbf{9.87} & \textbf{7.08} \\ 
\rowcolor{gray!25}\; \quad + Ours w DTD & 131M & 240 &  \underline{10.20} & \underline{7.31} \\
\arrayrulecolor{black!50}
\cmidrule{1-5}
\arrayrulecolor{black}
\; MeanFlow-M/2 & 308M & 240 & 5.01 & 4.61 \\
\rowcolor{gray!25}\; \quad + Ours w MinSNR  & 308M & 240 & \underline{4.61} & \underline{4.30} \\ 
\rowcolor{gray!25}\; \quad + Ours w DTD & 308M & 240 & \textbf{4.43} & \textbf{4.10} \\
\arrayrulecolor{black!50}
\cmidrule{1-5}
\arrayrulecolor{black}
\; MeanFlow-L/2 & 459M & 240 & 3.84 & 3.35 \\
\rowcolor{gray!25}\; \quad + Ours w MinSNR  & 459M & 240 & \underline{3.79} & \underline{3.31} \\ 
\rowcolor{gray!25}\; \quad + Ours w DTD & 459M & 240 & \textbf{3.47} & \textbf{3.24} \\
\bottomrule \addlinespace[2pt]
\multicolumn{5}{l}{\textit{\textbf{Few-step diffusion/flow models from end-to-end training}}} \\\addlinespace[2pt]
\; iCT-XL/2$^\ddagger$~\cite{song2023improved} & 675M & - & 34.24 & 20.30 \\
\; Shortcut-XL/2~\cite{frans2025one} & 675M & 160 & 10.60 & -  \\
\; iMM-XL/2$^\dagger$~\cite{zhou2025inductive} & 675M & 3840 & -  & 7.77 \\
\; MeanFlow-XL/2+~\cite{geng2025mean} & 676M & 1000 & - & 2.20 \\
\arrayrulecolor{black!50}
\cmidrule{1-5}
\arrayrulecolor{black}
\; MeanFlow-XL/2 & 676M & 240 & 3.43 & 2.93 \\
\rowcolor{gray!25}\; \quad + Ours w DTD & 676M & 240 &  \textbf{2.87} & \textbf{2.64} \\
\bottomrule
\end{tabular}
}
\end{minipage}
\hfill
\begin{minipage}[t]{0.438\textwidth}
\centering
\resizebox{\linewidth}{!}{
\begin{tabular}{lcrc}
\toprule
Method &  \#Params & NFE & FID$\downarrow$ \\
\midrule
\multicolumn{3}{l}{\textit{\textbf{GANs / Normalizing Flows}}} \\ \addlinespace[2pt]
\; BigGAN~\cite{brock2018large} & 112M & 1 & 6.95\\
\; StyleGAN-XL~\cite{sauer2022stylegan} & 166M & 1 & 2.30 \\
\; GigaGAN~\cite{kang2023scaling} & 569M & 1 & 3.45 \\
\; STARFlow~\cite{gu2025starflow} & 1.4B & 1 & 2.40 \\ 
\midrule
\multicolumn{3}{l}{\textit{\textbf{Autoregressive / Masking models}}} \\ \addlinespace[2pt]
\; VQ-GAN~\cite{esser2021taming} & 227M & 1024 & 26.52 \\
\; MaskGIT~\cite{chang2022maskgit} & 227M &  8  & 6.18 \\
\; VAR~\cite{tian2024visual}    & 2B     & 10$\times$2  & 1.92 \\ 
\; MAR-H~\cite{li2024autoregressive}   & 943M   &  256$\times$2  & 1.55 \\ 
\midrule
\multicolumn{3}{l}{\textit{\textbf{Diffusion / Flow models}}} \\ \addlinespace[2pt]
\; ADM~\cite{dhariwal2021diffusion} &  554M  & 250$\times$2 & 10.94 \\
\; LDM~\cite{rombach2022high} & 400M & 250$\times$2 & 3.60 \\
\; U-ViT-H/2~\cite{bao2023all} & 501M & 50$\times$2 & 2.29 \\ 
\; SimDiff~\cite{hoogeboom2023simple} & 2B & 512$\times$2 & 2.77 \\ 
\; DTR-L/2~\cite{park2023denoising} & 458M & 250$\times$2 & 2.33\\ 
\; DiT-XL/2~\cite{peebles2023scalable} & 675M & 250$\times$2 & 2.27 \\ 
\; SiT-XL/2~\cite{ma2024sit} & 675M & 250$\times$2 & 2.06 \\
\bottomrule
\end{tabular}
}
\end{minipage}
\vspace{-0.3cm}
\caption{\textbf{Results for class-conditional generation on ImageNet 256$\times$256.}
\textit{(Left)} Comparison of few-step diffusion/flow models.
\textit{(Right)} Other generative model families as reference.
``$\dagger$'' in left table and ``$\times$2'' in right table indicate that Classifier-Free Guidance (CFG) doubles NFE per sampling step.
$\ddagger$: iCT results from~\cite{zhou2025inductive}.
}
\label{tab:imagenet_benchmark}
\end{table*}

% 2.6417858730486614 -> 0.25

% 0.2 -> 4.48

%% file: sec/6_experiments.tex
\section{Experiments}
\label{sec:experiments}

\input{figures/convergence_compare_graph}
\input{figures/convergence_qualitative}

In the following, we show that our method accelerates convergence and attains higher final performance.
Together, these results show that our observations translate into practical training improvements.

\subsection{Experimental Setup}

%\paragraph{Setups.}
We follow the original experimental setup of MeanFlow~\cite{geng2025mean} and conduct experiments on ImageNet~\cite{deng2009imagenet} generation at 256$\times$256 resolution with DiT architectures~\cite{peebles2023scalable}. 
To measure few-step generation performance, we utilize the FID~\cite{heusel2017gans} score on 50K samples from either 1-NFE or 2-NFE generation.

\vspace{-0.4cm}
\paragraph{Implementation details.}
We test several ways of accelerating velocity training, one method from each category. For their simplicity and good empirical performance, we choose MinSNR~\cite{hang2023efficient} as loss weighting approach, and DTD~\cite{kim2024denoising} as timestep sampling approach.
More sophisticated methods exist~\cite{go2023addressing, wang2025closer, zheng2025bidirectional}, but we do not expect large differences in the context of our progressive weighting scheme.

%%% bookmark Konrad

\subsection{Comparative Evaluation}
\label{sec:comparative_study}

\paragraph{Comparison to MeanFlow.}
As shown in Table~\ref{tab:imagenet_benchmark} (top left), we first compare our method against MeanFlow using the DiT-B/4, DiT-M/2, and DiT-L/2 models.
Our method consistently outperforms MeanFlow, regardless of the velocity learning acceleration technique employed.
When comparing the two acceleration methods, MinSNR and DTD, a clear pattern emerges: MinSNR surpasses DTD on DiT-B/4, but this advantage diminishes at larger scales (L/2, M/2). 
We attribute this discrepancy to MeanFlow's adaptive loss weighting, which normalizes loss values by their norm to balance their influence. 
Loss-weighting strategies like MinSNR interfere with this adaptive mechanism, consequently reducing robustness across different model scales.
In contrast, timestep-sampling methods like DTD only modify the sampling distribution and leave the loss weighting scheme intact, thus preserving compatibility with MeanFlow's adaptive design.
Therefore, due to its consistent performance across model sizes, we select DTD as our primary acceleration method.

\vspace{-0.4cm}
\paragraph{ImageNet 256$\times$256 benchmark.}
We scale up our method with DTD to the DiT-XL model and compare it with previous one- and few-step diffusion/flow models in Table~\ref{tab:imagenet_benchmark} (bottom left).
Our method also outperforms MeanFlow in the DiT-XL setup (240 epochs), improving MeanFlow's 1-NFE FID from 3.43 to 2.87 and its 2-NFE FID from 2.93 to 2.64.
We highlight that this improvement on DiT-XL (3.43→2.87), a notable $16\%$ reduction in FID, substantially narrows the gap between one-step generation and multi-step diffusion models, as shown in Table~\ref{tab:imagenet_benchmark} (right).

\vspace{-0.3cm}
\paragraph{Convergence speed.}
Next, we compare the convergence behavior of our method against vanilla MeanFlow training, both quantitatively and qualitatively.
As shown in Fig.~\ref{fig:teaser} and~\ref{fig:convergence_graph}, our approach achieves substantially faster convergence across all model sizes.
Notably, our method with DTD demonstrates superior convergence on DiT-XL/2, -L/2, and -M/2, achieving approximately 2.5$\times$, 2.3$\times$, and 2.1$\times$ speedup, respectively.
Qualitatively, Fig.~\ref{fig:convergence_qualitative} shows samples from DiT-XL/2 at different training epochs.
At equivalent epochs, our method generates noticeably sharper and more detailed images than vanilla MeanFlow.
Remarkably, our samples at 120 epochs exhibit quality comparable to or exceeding vanilla MeanFlow's 240-epoch results, demonstrating $\approx$2$\times$ faster convergence.
Together, these quantitative and qualitative results confirm that our approach not only substantially accelerates training but also improves final generation quality.

\input{tables/cfg_configuration}

\subsection{Ablation and Analysis}

\paragraph{Effectiveness across different CFG configurations.}
MeanFlow integrates Classifier-Free Guidance (CFG) by mixing three components in the target $u_{\text{tgt}}$: conditional velocity, conditional velocity prediction, and unconditional velocity prediction, using coefficients $\omega$ and $\kappa$ (see Appendix A in~\cite{geng2025mean}).
For smaller models (DiT-B and DiT-M), $\kappa=0.5$ and $\omega=2.0$ are used, while larger models (DiT-L and DiT-XL) use $\kappa=0.92$ and $\omega=2.5$.
Additionally, for DiT-L and DiT-XL, CFG mixing is only applied when $t$ is sampled within specific ranges ([0.0, 0.8] for DiT-L, [0.0, 0.75] for DiT-XL).
To verify our method's robustness to these configurations, we train DiT-L and DiT-XL using the CFG settings of both the small and large models.

Table~\ref{tab:cfg_configuration} shows that our method consistently improves performance across all CFG configurations. 
This demonstrates the robustness of our method to different guidance scenarios, underscoring its role as a simple but highly effective technique for achieving consistent performance gains.

\vspace{-0.4cm}
\paragraph{Effect of individual components.}
To validate the contribution of each component, we train DiT-B/4 with three configurations: (1) applying only velocity acceleration methods (MinSNR or DTD), (2) applying only progressive $\mathcal{L}_u$ weighting, and (3) the full method combining both components.
All models are trained for 240 epochs, matching the setup in Sec.~\ref{sec:comparative_study}.
The results are illustrated in Table~\ref{tab:abl_each_component}.
As shown in the results, only applying velocity acceleration methods improves vanilla MeanFlow training (11.58 → 10.57 with MinSNR, 10.96 with DTD at 1-NFE), demonstrating that rapid $v$ formulation benefits training.
Moreover, applying only progressive weighting on $\mathcal{L}_u$ improves the performance to 10.98 at 1-NFE, showing the benefit of proper temporal gap scheduling.
The combined approach yields the strongest results, with a combination of MinSNR and weighting yielding an FID of 9.87 at 1-NFE and 10.20 when combining DTD and weighting.
This demonstrates that the two components are complementary: velocity acceleration rapidly establishes the instantaneous velocity foundation, while progressive weighting drives effective average velocity learning.

\input{tables/component_abl}

\input{tables/effect_k}

\vspace{-0.4cm}
\paragraph{Effect of schedule parameter $k$.}
While the pace of our progressive weighting $s = 1 - (i/T)^k$ can be modulated by adjusting $k$, we use $k=1$ (linear schedule) in all experiments for simplicity.
To verify the impact of this choice, we compare different $k$ values on DiT-B/4 in Table~\ref{tab:abl_effect_k}.
The results show that $k=1$ achieves the best performance (10.20 FID), validating our choice of a simple linear schedule.
We observe that slower transitions ($k < 1$) and faster transitions ($k > 1$) both degrade performance, indicating that a linear schedule provides a good balance.

\input{tables/velocity_analysis}

\vspace{-0.3cm}
\paragraph{Enhanced instantaneous velocity.}

We argue that accelerating instantaneous velocity formation is crucial for effective training (Sec.~\ref{sec:observation}).
To verify that the acceleration methods improve $v$ quality, we evaluate multi-step generation performance using $u_\theta(z_t, t, t)$ as the velocity estimate (which equals the model's instantaneous velocity prediction when $r=t$).
As shown in Table~\ref{tab:multi_step}, our methods consistently outperform MeanFlow across all sampling steps (32, 64, 128 NFE) regardless of the applied velocity acceleration methods. 
This demonstrates that our training strategy actually improves the underlying instantaneous velocity estimated by the model, confirming that our approach yields higher-quality velocity fields.

% \paragraph{Convergence analysis of $v$.}

% \begin{itemize}
%     \item 1) How do MinSNR and DTD applied training fastly converge $v$?
%     \item 2) Is there any correlation between convergence trend $v$ and $u$?
% \end{itemize}

% CIFAR-10 -> Supplementary
% FFHQ -> Supplementary.
% - Longer Training. -> For iteration generability (schedule generalizability) -> Supplementary
% - 512x512 setup.

% Considered points
% Distillation setup? -> Show generalizability of our method.
% FFHQ -> Good point. We have to show our method works well in other dataset.

% v vs small-\Delta learning comparison of curvature.
% Small-Δt supervision effectively constrains learning to locally linear regions of the flow manifold, where curvature is low and the dynamics are smooth.
% Once those local dynamics are well captured, the model can progressively handle higher-curvature (nonlinear) regions as Δt increases.
% In other words, early small-Δt training flattens the curvature landscape the model has to integrate over later.

%% file: figures/convergence_compare_graph.tex
\begin{figure*}[t!]
    \centering
    % lenend figure
    \begin{subfigure}[t]{0.7\linewidth}
        \centering
        \includegraphics[width=\linewidth]{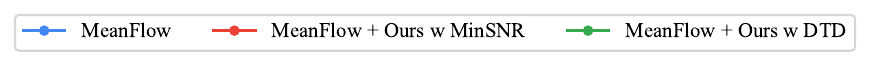}
    \end{subfigure} \\
    \vspace{-0.2cm}
    \begin{subfigure}[t]{0.33\linewidth}
        \centering
        \includegraphics[width=\linewidth]{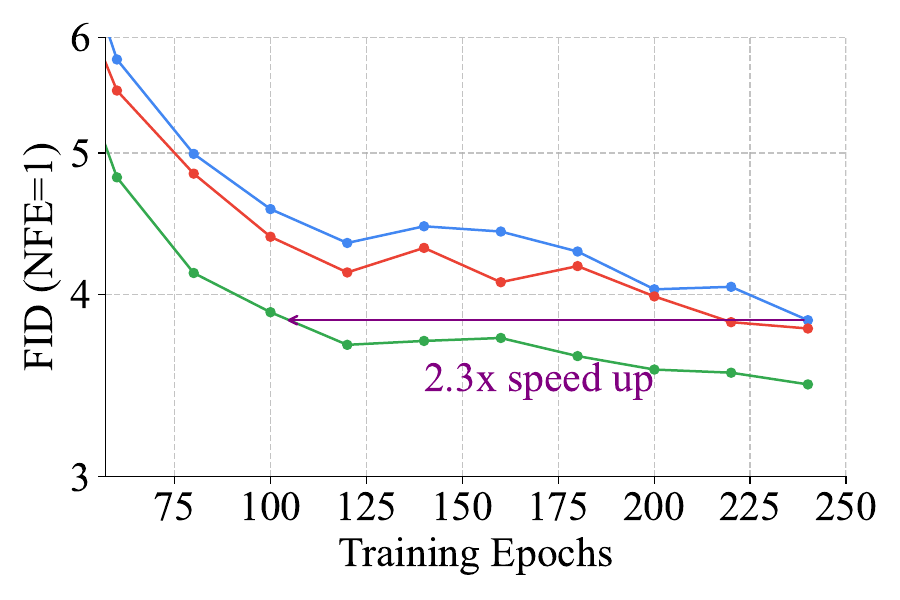}
        \vspace{-0.6cm}
        \caption{DiT-L/2}
    \end{subfigure}
    \hfill
    \begin{subfigure}[t]{0.33\linewidth}
        \centering
        \includegraphics[width=\linewidth]{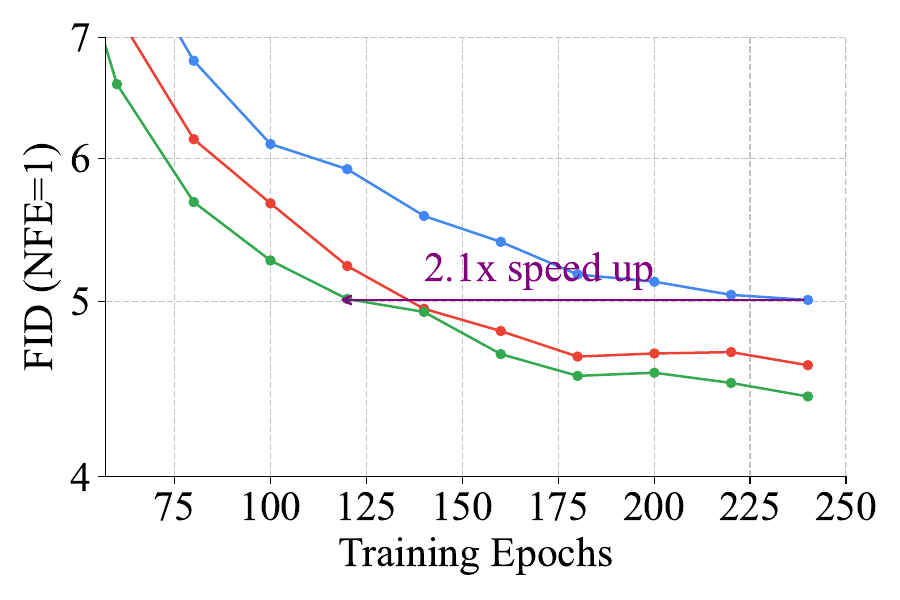}
        \vspace{-0.6cm}
        \caption{DiT-M/2}
    \end{subfigure}
    \hfill
    \begin{subfigure}[t]{0.33\linewidth}
        \centering
        \includegraphics[width=\linewidth]{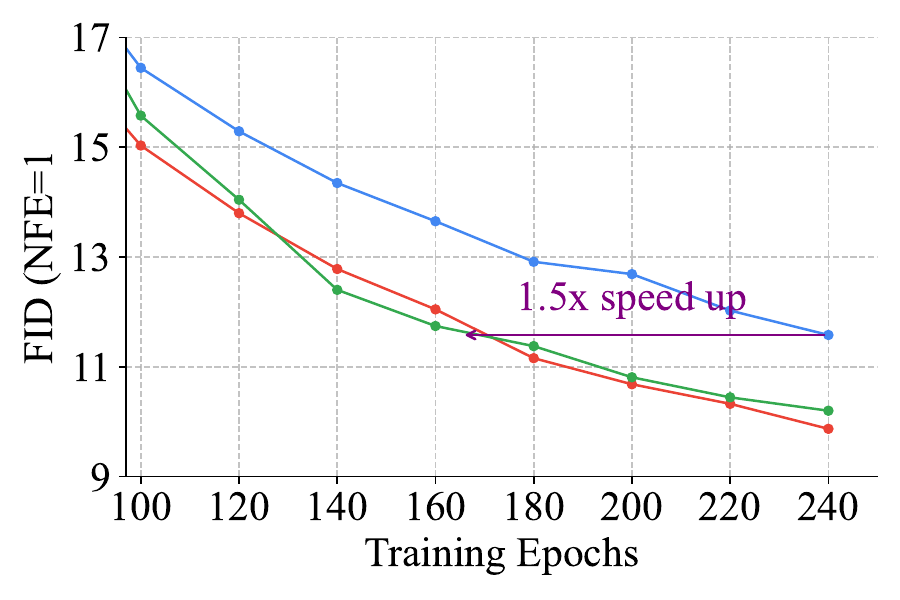}
        \vspace{-0.6cm}
        \caption{DiT-B/4}
    \end{subfigure}
    \vspace{-0.5cm}
    \caption{\textbf{Convergence speed comparison between MeanFlow and our methods across model sizes.}}
    \label{fig:convergence_graph}
\end{figure*}

%% file: figures/convergence_qualitative.tex
\begin{figure*}[t!]
    \centering
    \includegraphics[width=1\textwidth]{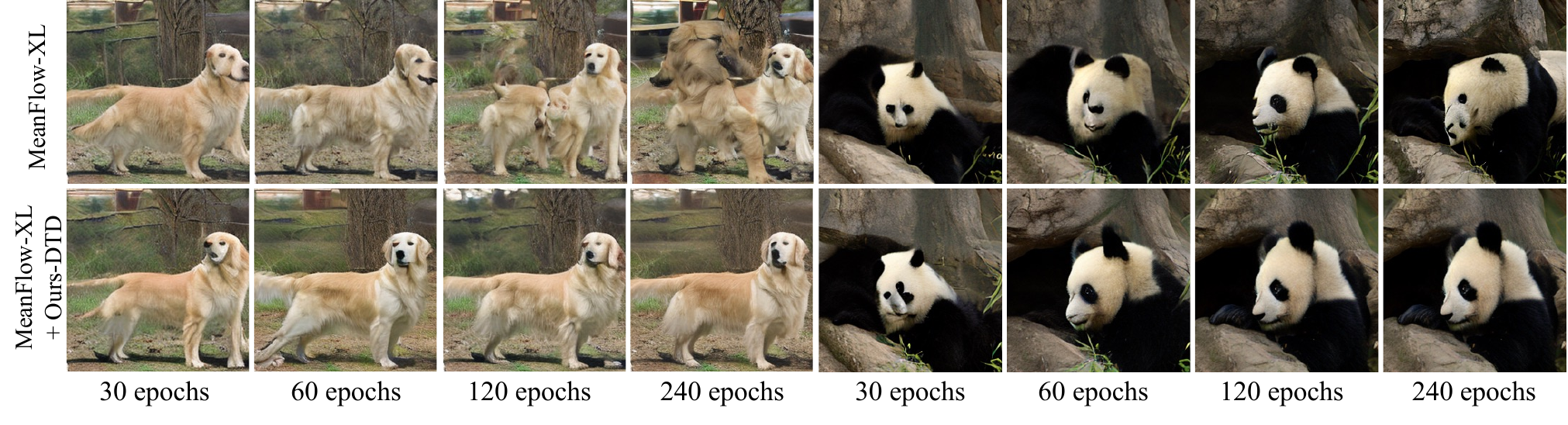}
    \vspace{-0.7cm}
    \caption{\textbf{Qualitative comparison of generated samples across training epochs on DiT-XL/2.}
    }
    \label{fig:convergence_qualitative}
\end{figure*}

%% file: tables/cfg_configuration.tex
\begin{table}[t]
    \centering
    \resizebox{\linewidth}{!}{
    \begin{tabular}{c|lcc}
        \toprule
         CFG Setup & Method & FID (1-NFE)$\downarrow$ & FID (2-NFE)$\downarrow$ \\
         \midrule
         & MeanFlow-L/2 & 4.60 & 4.58 \\ 
         $\kappa=0.5$ & \cellcolor{gray!25} \quad + Ours w DTD &\cellcolor{gray!25} \textbf{4.08} & \cellcolor{gray!25} \textbf{3.97} \\ 
         $\omega=2.0$ & MeanFlow-XL/2 & 4.72 & 4.64 \\ 
         & \cellcolor{gray!25} \quad + Ours w DTD  &  \cellcolor{gray!25} \underline{4.31} & \cellcolor{gray!25} \underline{4.29} \\ 
         \midrule
         $\kappa=0.92$ & MeanFlow-L/2 & 3.84 & 3.35 \\ 
         $\omega=2.5$ & \cellcolor{gray!25} \quad + Ours w DTD & \cellcolor{gray!25} \textbf{3.47} & \cellcolor{gray!25} \textbf{3.24} \\
         \midrule
         $\kappa=0.92$ & MeanFlow-XL/2 & 3.43 & 2.93 \\ 
         $\omega=2.5$ & \cellcolor{gray!25} \quad + Ours w DTD &\cellcolor{gray!25} \textbf{2.87} & \cellcolor{gray!25} \textbf{2.64} \\
         \bottomrule
    \end{tabular}}
    \vspace{-0.3cm}
    \caption{
    \textbf{Robustness to CFG configurations.}
    Performance of DiT-L and DiT-XL with different CFG setups.
    Our method consistently improves across all configurations.
    }
    \label{tab:cfg_configuration}
\end{table}

%% file: tables/component_abl.tex
\begin{table}[t]
    \centering
    \resizebox{\linewidth}{!}{
    % \smapp
    \begin{tabular}{l|cc}
        \toprule
        Method & FID (1-NFE)$\downarrow$ & FID (2-NFE)$\downarrow$ \\
         \midrule
         MeanFlow-B/4 & 11.58 & 7.85 \\ \midrule
         \; + MinSNR & 10.57 & 7.38 \\  
         \; + DTD & 10.96 & 7.55 \\ \midrule
         \; + $\mathcal{L}_u$ weighting. & 10.98 & 7.58 \\ 
        \midrule
        \rowcolor{gray!25} \; + MinSNR + $\mathcal{L}_u$ weighting. & \textbf{9.87} & \textbf{7.08} \\
         \rowcolor{gray!25} \; + DTD + $\mathcal{L}_u$ weighting. & \underline{10.20} & \underline{7.31} \\  \bottomrule
    \end{tabular}}
    \vspace{-0.3cm}
    \caption{
    \textbf{Ablation of method components.}
    Velocity acceleration methods and $\mathcal{L}_u$ weighting each improve upon vanilla MeanFlow training, with their combination achieving the best performance.
    }
    \label{tab:abl_each_component}
\end{table}

%% file: tables/effect_k.tex
\begin{table}[t]
    \centering
    \resizebox{0.99\linewidth}{!}{
    \begin{tabular}{lccccc}
        \toprule
         & \multirow{2}{*}{MeanFlow-B/4} & \multicolumn{4}{c}{MeanFlow-B/4 + Ours w DTD}\\ \cmidrule{3-6} 
         & & $k=0.5$ &$k=1$ & $k=2$ & $k=3$ \\
         \midrule
         FID$\downarrow$ (1-NFE) & 11.58 & \cellcolor{gray!25} \underline{11.16} & \cellcolor{gray!25} \textbf{10.20} & \cellcolor{gray!25} 11.44 & \cellcolor{gray!25} 11.99 \\ \midrule
    \end{tabular}
    }
    \vspace{-0.3cm}
    \caption{%\textbf{Effect of $k$.} 
    \textbf{1-NFE FID with varying settings of the schedule parameter $k$}. 
    The linear schedule ($k=1$) achieves the lowest FID.
    }
    \vspace{-0.3cm}
    \label{tab:abl_effect_k}
\end{table}

%% file: tables/velocity_analysis.tex
\begin{table}[t]
\centering
\small
\resizebox{0.99\linewidth}{!}{
\begin{tabular}{lccc}
\toprule
\multirow{2}{*}{Method} & \multicolumn{3}{c}{FID$\downarrow$} \\ \cmidrule{2-4}
& 32-NFE & 64-NFE & 128-NFE \\ 
\midrule
MeanFlow & 7.61  & 7.26 & 7.16 \\
\rowcolor{gray!25} \;  + Ours w MinSNR & \textbf{7.09} & \textbf{6.91} & \textbf{6.86} \\
\rowcolor{gray!25} \;  + Ours w DTD & \underline{7.25} & \underline{7.01} & \underline{6.93} \\
\bottomrule
\end{tabular}
}
\vspace{-0.2cm}
\caption{\textbf{Comparison of multi-step generation with estimated instantaneous velocity on DiT-B/4 models.}}
\vspace{-0.3cm}
\label{tab:multi_step}
\end{table}

%% file: sec/7_conclusion.tex
\section{Conclusion}
\label{sec:conclusion}

In this work, we revisited MeanFlow through a detailed analysis of its instantaneous and average velocity components and their interaction during training. 
We identified key learning dynamics: instantaneous velocity should be established early to enable effective average velocity learning, and supervision on small temporal gaps creates a more favorable foundation for subsequently learning large-gap average velocity.
Building on these insights, we propose an improved training strategy that accelerates instantaneous velocity formation and prioritizes small temporal gaps in the early training phase before gradually transitioning to larger gaps.
Our approach achieves faster convergence, improved one-step generation, and reduced training cost. 
We believe these findings offer both a deeper understanding of MeanFlow’s learning behavior and a practical foundation for designing efficient few-step generative models.
\paragraph{Acknowledgements.}
This work was supported by ETHAR (ETH Zürich Augmented Reality Research Lab), funded through a Google donation; and by a grant from the Swiss National Supercomputing Center within the Swiss AI Initiative (project a144). Julius Erbach is supported by the Max Planck ETH Center for Learning Systems.

%% file: sec/X_suppl.tex
\clearpage
\setcounter{page}{1}
\maketitlesupplementary

\appendix

\section{Setups for Observational Study}

In this section, we provide additional details for the observational study presented in Sec.~\textcolor{cvprblue}{4}.

\paragraph{Impact of $v$-learning on $u$-learning.}
For the $v$-pretraining stage, we set the flow matching ratio (FM ratio) as $\text{FM ratio}=100\%$, \ie, we always sample transitions with $t = r$. Except for the FM ratio (the probability of sampling $t = r$) and the number of pretraining epochs, all hyperparameters (optimizer, learning rate schedule, batch size, network architecture, and regularization) are kept identical to those used in the main training in Sec.~\textcolor{cvprblue}{6}.

\paragraph{Impact of $u$-learning on $v$-learning.}
For the training setup where we explicitly control the temporal gap $\Delta t$, we set the sampling ratio of $t = r$ to zero.
Given a randomly sampled $(t, r)$, we first compute the raw gap $\Delta t_{\text{raw}} = t - r$ and then rescale $\Delta t_{\text{raw}}$ into a prescribed target range. Using the rescaled gap ${\Delta \tilde{t}}$ together with the originally sampled $t$, we redefine the reward time as $r := t - {\Delta \tilde{t}}$.

\paragraph{Task affinity analysis.}
The task affinity score is defined as the cosine similarity between the gradients of the two tasks across training iterations. 
Concretely, at the end of each training epoch, we randomly sample 5K data points and compute the gradients of the $v$-loss and the $u$-loss with 4 intervals described in Fig~\textcolor{cvprblue}{5}, respectively.
We then compute the gradient cosine similarity within each temporal interval of the losses and average these values over all samples and training epochs to obtain the final task-affinity score.

\section{Detailed Integration into MeanFlow}

\paragraph{Recap of MeanFlow.}
To recall, the MeanFlow~\cite{geng2025mean} objective in Eq.~\ref{eq:loss} reduces to standard flow matching by learning the instantaneous velocity $v$ if $t=r$. It can be decomposed into two components, one for the average velocity $\mathcal{L}_u(z_t, r, t)$ and the other for the instantaneous velocity $\mathcal{L}_v(z_t, t)$:
\begin{equation}
    \label{appeq:meanflow_remind}
    \mathcal{L}_\text{MF}
    =
    \mathbb{E}_{x, \epsilon, t, r}
    \big[
        \mathcal{L}_u(z_t, r, t)\,\mathbb{I}(t \neq r)
        +
        \mathcal{L}_v(z_t, t)\,\mathbb{I}(t = r)
    \big],
\end{equation}
where $\mathbb{I}(\cdot)$ is the indicator function, $\mathcal{L}_u$ is applied only when $t$ does not coincide with $r$, and $\mathcal{L}_v$ is applied when $t = r$.

Furthermore, MeanFlow introduces additional techniques for improving its training dynamics.
Among these techniques, \emph{adaptive loss weighting} deserves particular attention, especially since our method also employs specific weighting strategies. This mechanism rescales the loss according to the magnitude of the regression error.
Specifically, let $e$ denote the regression error vector, defined as $u_\theta(z_t, t, t) - v_t(z_t|\epsilon)$ for the instantaneous velocity term $\mathcal{L}_v$, and $u_\theta(z_t, r, t) - \text{sg}(u_{\text{tgt}})$ for the average velocity term $\mathcal{L}_u$.
MeanFlow calculates an adaptive weight $w_{\text{adp}}$ to normalize the loss scale:
\begin{equation}
    w_{\text{adp}} = \text{sg}\left(\frac{1}{(\| e \|_2^2 + c)^p}\right),
\end{equation}
where $c$ is a small constant for numerical stability and $p$ controls the normalization strength. 
The final per-sample loss is then computed by multiplying this adaptive weight by the squared $L_2$ norm of the regression error. 
Empirically, setting $p=1$ in MeanFlow has been demonstrated to work effectively. Therefore, we adopt this setting in our loss formulation.

\paragraph{Integration of our method.}
Let $\mathcal{L}_v^{\text{adp}}(z_t, t)$ and $\mathcal{L}_u^{\text{adp}}(z_t, r, t)$ denote the per-sample loss terms for the instantaneous and average velocities, respectively.
The adaptive normalization would be ineffective if our proposed weighting schedules—specifically $\alpha(t)$ for velocity learning and $\beta(\Delta t, s)$ for progressive training— were applied directly to the raw losses.
Therefore, we strictly apply these weightings \emph{after} the adaptive normalization.

In particular, when we integrate our method with a loss-weighting method (\eg, MinSNR~\cite{hang2023efficient}), the objective function is formulated as:
\begin{equation}
    \label{appeq:ours_weighting}
    \begin{split}
        \mathcal{L}_\text{ours}^\text{weighting}
        = \mathbb{E}_{x, \epsilon, t, r}
        \big[
            &\beta(\Delta t, s) \cdot \mathcal{L}_u^{\text{adp}}(z_t, r, t) \cdot \mathbb{I}(t \neq r) \\
            &+ \alpha(t) \cdot \mathcal{L}_v^{\text{adp}}(z_t, t) \cdot \mathbb{I}(t = r)
        \big].
    \end{split}
\end{equation}

Alternatively, when integrating timestep sampling techniques (\eg, DTD~\cite{kim2024denoising}), where the sampling distribution is modified to $p_{\text{acc}}(t)$ instead of applying explicit loss weighting on $v$, the objective becomes:
\begin{equation}
    \label{appeq:ours_sampling}
    \begin{split}
        \mathcal{L}_\text{ours}^\text{sampling}
        = \mathbb{E}_{x, \epsilon, t \sim p_{\text{acc}}(t), r}
        \big[
            &\beta(\Delta t, s) \cdot \mathcal{L}_u^{\text{adp}}(z_t, r, t) \cdot \mathbb{I}(t \neq r) \\
            &+ \mathcal{L}_v^{\text{adp}}(z_t, t) \cdot \mathbb{I}(t = r)
        \big].
    \end{split}
\end{equation}

\section{Additional Experimental Results}

\subsection{Observational Study on FFHQ Dataset}

While our primary analysis was conducted on ImageNet, we verify whether these findings generalize to unconditional image generation on the FFHQ dataset in this section. 
Figures~\ref{appendix_fig:observation_1}--\ref{appendix_fig:observation_3} demonstrate that our key observations consistently hold on FFHQ.

\input{figures/appn_obs1_clue1}

\paragraph{Learning $v$ facilitates learning $u$.}
Figure~\ref{appendix_fig:observation_1} presents results corresponding to those in Fig.~\textcolor{cvprblue}{2} of the main paper.
Note that the experimental settings are slightly adjusted to accommodate the different datasets. Since the number of iterations per epoch decreases significantly from ImageNet (5,004 steps) to FFHQ (273 steps), we increase the number of training epochs to compensate for the reduced training volume.

In the first setting (Top), we fix the budget for $u$-finetuning at 70 epochs while varying the duration of the $v$-pretraining across $\{0, 100, 110, 120, 130\}$ epochs.
In the second setting (Bottom), we fix the total training budget at 200 epochs and allocate $\{0, 100, 110, 120, 130\}$ epochs to $v$-pretraining, with the remaining epochs used for $u$-finetuning.

Consistent with our main results, the findings clearly indicate that a heavier investment in $v$-pretraining yields more stable and accurate $u$-learning.
In the fixed-budget scenario (Bottom), prioritizing $v$ in the early stages is more effective and thereby accelerates convergence.
Overall, these results reinforce the conclusion from the main paper that a well-formed instantaneous velocity $v$ is a necessary prerequisite for learning the average velocity $u$.

\input{figures/appn_obs1_clue2}

\paragraph{Corruption in $v$-learning disrupts $u$-learning.}
Figure~\ref{appendix_fig:observation_2} corresponds to Fig.~\textcolor{cvprblue}{3} in the main paper.
Consistent with the previous setup, we extended the training duration to 200 epochs for the FFHQ dataset.
We observe trends identical to those reported in Section~\textcolor{cvprblue}{4.1}: even with small noise ($k = 0.03$), $u$-learning severely degrades.
This confirms that a corrupted instantaneous velocity makes learning the average velocity substantially more difficult.

\paragraph{Impact of $u$-Learning on $v$-Learning.}

\input{figures/appn_obs2_clue1}

Figure~\ref{appendix_fig:observation_3} corresponds to Figure~\textcolor{cvprblue}{4} in the main text.
To adapt to the FFHQ dataset, we extended the training duration for both stages (pretraining and finetuning) from 40 to 150 epochs.
Specifically, we compare models starting from either random initialization or a model pretrained with $v$-loss for 150 epochs, followed by 150 epochs of $u$-finetuning restricted to specific $\Delta t$ ranges.

The findings align perfectly with those on ImageNet:
\begin{itemize}
    \item \textbf{Small $\Delta t$ supervision forms $v$:} Restricting $u$-learning to small temporal gaps ($\Delta t \in [0.1, 0.3]$) proves to be a viable proxy for $v$-learning, achieving 32-NFE FID scores comparable to pure $v$-pretraining (green line) when trained from scratch. Furthermore, it yields additional performance gains when applied to the $v$-pretrained model, indicating effective refinement of the instantaneous velocity.
    \item \textbf{Large $\Delta t$ supervision deteriorates $v$.} In contrast, supervision with larger temporal gaps ($\Delta t > 0.3$) results in poor 32-NFE FID across both initializations. 
    This confirms that large-$\Delta t$ supervision fails to establish $v$ when training from scratch and also disrupts an already well-formed velocity field.
\end{itemize}

\subsection{Ablation of Individual Components on FFHQ dataset}

\input{tables/appn_component_abl}

To further demonstrate and analyze the effectiveness of our method, we conducted an additional ablation study on the FFHQ dataset, mirroring the experimental design of Table~\textcolor{cvprblue}{3} in the main paper.
For this experiment, we utilized the DiT-B/2~\cite{peebles2023scalable} architecture and trained all models for 400 epochs.
The results are summarized in Table~\ref{apptab:abl_each_component}.

Consistent with the results on ImageNet, applying velocity acceleration methods alone improves upon vanilla MeanFlow training ($12.90 \to 12.41$ with DTD at 1-NFE), demonstrating that rapid formulation of $v$ benefits training performance.
However, the improvement from MinSNR is relatively marginal ($12.90 \to 12.82$).
As discussed in Section~\ref{sec:comparative_study}, this is likely because explicit loss weighting strategies interfere with the adaptive loss weighting mechanism inherent to MeanFlow, consequently reducing robustness across different model scales.
Moreover, applying only progressive weighting on $\mathcal{L}_u$ improves the performance to $12.24$ at 1-NFE, demonstrating the benefit of proper temporal gap scheduling.
The combined approach yields the strongest results, achieving an FID of 11.33 at 1-NFE and 9.19 at 2-NFE.

%% file: figures/appn_obs1_clue1.tex
\begin{figure}[t]
    \centering
    % lenend figure
    \begin{subfigure}[t]{0.9\linewidth}
        \centering
        \includegraphics[width=\linewidth]{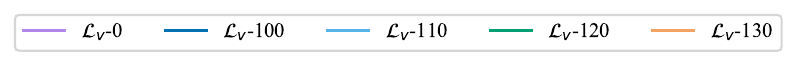}
        % \caption{\textcolor{blue}{HJ: is subfigure caption needed?}}
    \end{subfigure}
    \vspace{-5pt}
    \begin{subfigure}[t]{0.99\linewidth}
        \centering
        \includegraphics[width=\linewidth]{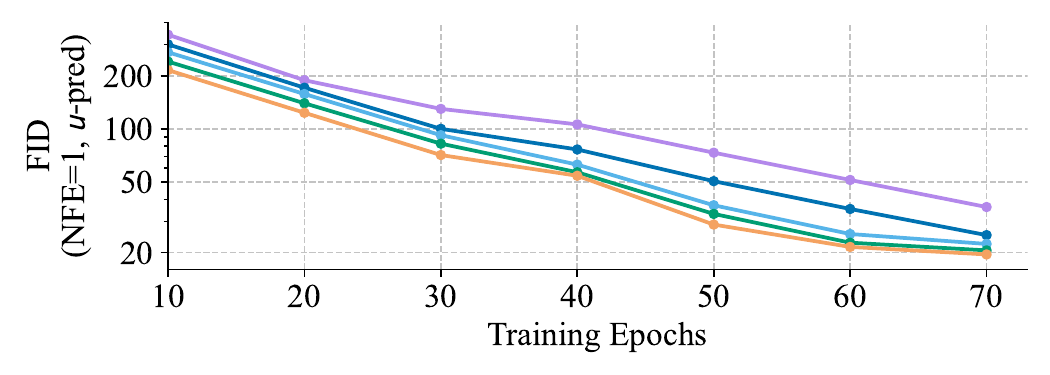}
        % \caption{\textcolor{blue}{HJ: is subfigure caption needed?}}
    \end{subfigure}
    \vspace{-5pt}
    \begin{subfigure}[t]{0.98\linewidth}
        \centering
        \includegraphics[width=\linewidth]{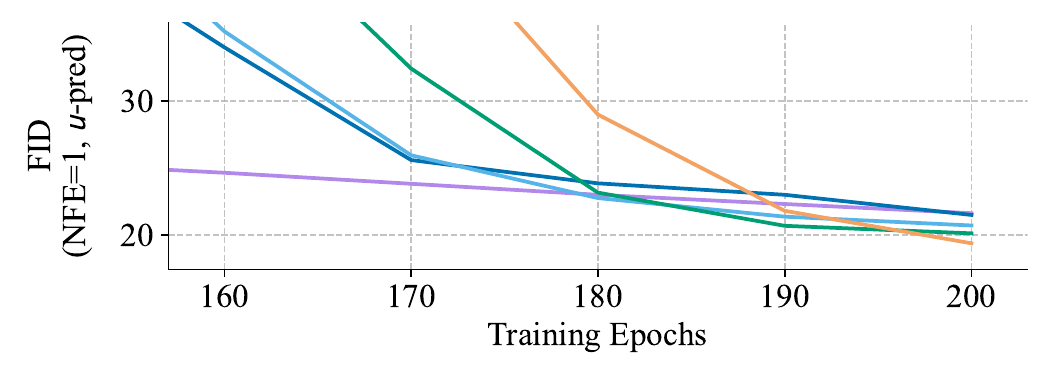}
        % \caption{\textcolor{blue}{HJ: is subfigure caption needed?}}
    \end{subfigure}
    \vspace{-0.2cm}
    \caption{\textbf{$v$-learning facilitates $u$-learning.} \emph{(Top)} 1-NFE FID during $u$-finetuning according to $v$-pretraining epochs.
    \emph{(Bottom)} 1-NFE FID under a fixed 200-epoch budget with varying allocation between $v$-pretraining and $u$-finetuning.
    Both settings show that investing in $v$-learning improves $u$-learning quality.
    }
    \vspace{-0.2cm}
    \label{appendix_fig:observation_1}
\end{figure}

%% file: figures/appn_obs1_clue2.tex
\begin{figure}
    \centering
    \includegraphics[width=\linewidth]{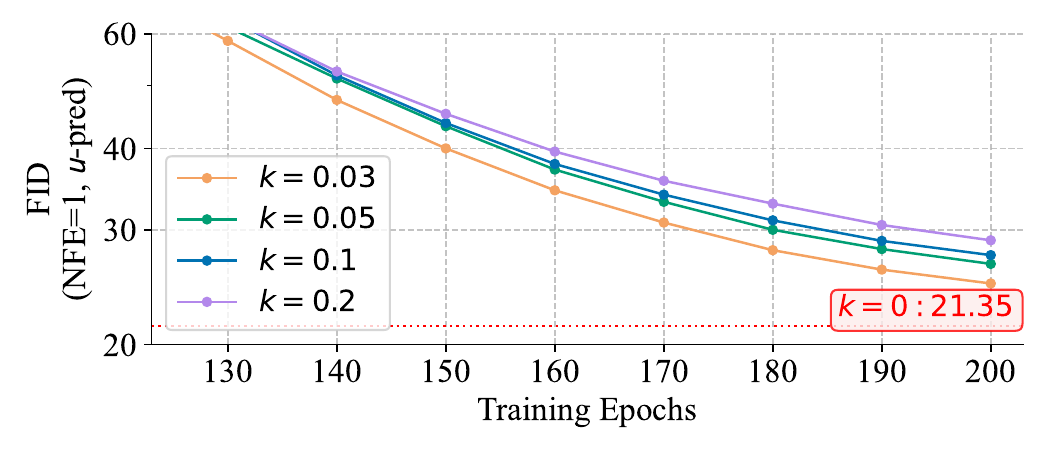}
    \vspace{-0.6cm}
    \caption{
    \textbf{Corruption in $v$-learning disrupts $u$-learning.} 
    1-NFE FID when training with $\mathcal{L}_{\mathrm{MF}}$ while injecting Gaussian noise scaled by $k\!\cdot\!\|v_t(z_t|\epsilon)\|$ into the target velocity of $\mathcal{L}_v$. 
    Even small noise ($k = 0.03$) disrupts $v$-learning and severely degrades $u$-learning performance compared to clean training ($k=0$).
    }
    \vspace{-0.2cm}
    \label{appendix_fig:observation_2}
\end{figure}

%% file: figures/appn_obs2_clue1.tex
\begin{figure}
    \centering
    \includegraphics[width=1\linewidth]{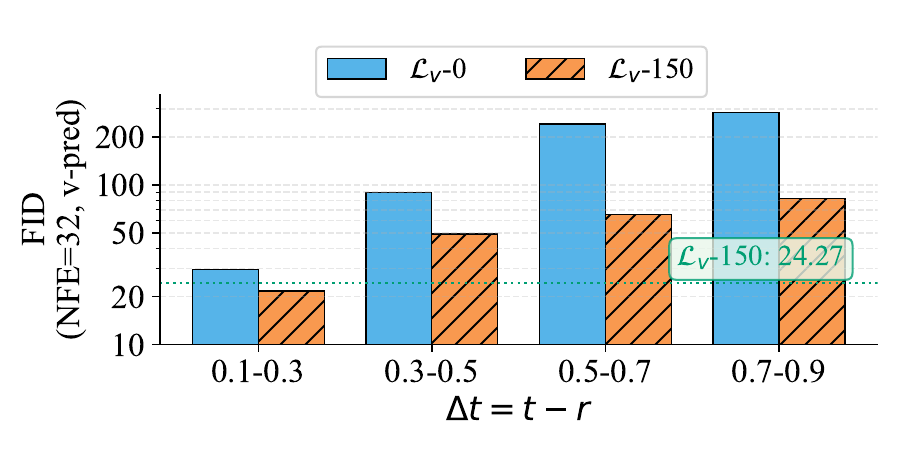}
    \vspace{-0.9cm}
    \caption{
    \textbf{Impact of $\Delta t$ of $u$-learning on $v$-learning.}
    32-NFE FID after 150 epochs of $u$ finetuning across different $\Delta t$ ranges, starting from either random initialization (blue) or $v$-pretrained model (orange, 150 epochs).
    Small $\Delta t$ enables constructing and improving $v$, while large $\Delta t$ degrades pretrained $v$. The green line denotes the performance of the $v$-pretrained model.
    }
    \vspace{-0.2cm}
    \label{appendix_fig:observation_3}
\end{figure}

%% file: tables/appn_component_abl.tex
\begin{table}[t]
    \centering
    \resizebox{\linewidth}{!}{
    % \smapp
    \begin{tabular}{l|cc}
        \toprule
        Method & FID (1-NFE)$\downarrow$ & FID (2-NFE)$\downarrow$ \\
         \midrule
         MeanFlow-B/2 & 12.90 & 9.81 \\ \midrule
         \; + MinSNR & 12.82 & 9.65 \\  
         \; + DTD & 12.41 & 9.61 \\ \midrule
         \; + $\mathcal{L}_u$ weighting. & 12.24 & 9.52 \\ 
        \midrule
         \rowcolor{gray!25} \; + DTD + $\mathcal{L}_u$ weighting. & \textbf{11.33} & \textbf{9.19} \\  \bottomrule
    \end{tabular}
    }
    \vspace{-0.3cm}
    \caption{
    \textbf{Ablation of method components.}
    Velocity acceleration methods and $\mathcal{L}_u$ weighting each improve upon vanilla MeanFlow training, with their combination achieving the best performance. 
    }
    \label{apptab:abl_each_component}
\end{table}